\documentclass[letterpaper, 10 pt, conference]{ieeeconf}  

\IEEEoverridecommandlockouts                              

\usepackage{graphics} 
\usepackage{epsfig} 
\usepackage{mathptmx} 
\usepackage{times} 
\usepackage{amsmath} 
\usepackage{amssymb}  
\usepackage{svg}
\usepackage{url}
\usepackage{hyperref}
\usepackage{algorithm}
\usepackage{algpseudocode}
\let\labelindent\relax

\usepackage{enumitem}
\usepackage{xcolor}
\usepackage{cite}
\usepackage{array,colortbl,multirow,multicol,booktabs,ctable}
\title{\LARGE \bf
Optimal Driver Warning Generation in Dynamic Driving Environment
}

\author{ Chenran Li$^{1}$, Aolin Xu$^{2}$, Enna Sachdeva$^{2}$, Teruhisa Misu$^{2}$, and Behzad Dariush$^{2}$ 
\thanks{$^{1}$ Chenran Li is with Department of Mechanical Engineering, University of California, Berkeley, California, USA. The work was conducted during Chenran Li's internship at Honda Research Institute USA. {\tt\footnotesize chenran\_li@berkeley.edu}}%
\thanks{$^{2}$ Aolin Xu, Enna Sachdeva, Teruhisa Misu, and Behzad Dariush are with Honda Research Institute USA
        {\tt\small 
\{aolin\_xu, enna\_sachdeva, tmisu, bdariush\}@honda-ri.com}}%
}

\begin{document}

\maketitle
\thispagestyle{empty}
\pagestyle{empty}

\begin{abstract}
The driver warning system that alerts the human driver about potential risks during driving is a key feature of an advanced driver assistance system. Existing driver warning technologies, mainly the forward collision warning and unsafe lane change warning, can reduce the risk of collision caused by human errors. However, the current design methods have several major limitations. Firstly, the warnings are mainly generated in a one-shot manner without modeling the ego driver's reactions and surrounding objects, which reduces the flexibility and generality of the system over different scenarios. Additionally, the triggering conditions of warning are mostly rule-based threshold-checking given the current state, which lacks the prediction of the potential risk in a sufficiently long future horizon. In this work, we study the problem of optimally generating driver warnings by considering the interactions among the generated warning, the driver behavior, and the states of ego and surrounding vehicles on a long horizon. The warning generation problem is formulated as a partially observed Markov decision process (POMDP). An optimal warning generation framework is proposed as a solution to the proposed POMDP. The simulation experiments demonstrate the superiority of the proposed solution to the existing warning generation methods. 
\end{abstract}

\section{INTRODUCTION}
The road traffic plays an important role in people's lives. With the development of the complexity of city road networks, it is crucial for an advanced driver assistance system to be able to alert the potential risks to the human driver during driving. As shown in the studies of the human driver behavior with the warning system \cite{zhu2020impact,yue2021effects,kamrani2023drivers}, existing driver warning technologies, mainly the forward collision warning and unsafe lane change warning, can reduce the risk of collision caused by human errors. 

However, studies show that the human drivers' reactions to warnings vary with the type of warning and different drivers \cite{wu2021auditory,zhao2023effects}, while the existing methods have not addressed this phenomenon adequately. Most methods in the literature mainly generate the warning in a one-shot manner without modeling the ego driver's reactions and surrounding objects \cite{TSMC96, iranmanesh2018adaptive, YUAN202064}, which reduces the flexibility and generality of the system over different drivers and scenarios. Meanwhile, the triggering conditions of warning are mostly rule-based threshold-checking based on the current state, such as the time-to-collision (TTC) and the minimum safety distance \cite{pyo2016front, yang2020forward, guo2022forwarding}, which lacks the prediction of the potential risk in a sufficiently long future horizon. As a consequence, the current warning systems, while effective in preventing collisions, tend to prompt urgent and uncomfortable braking actions. Studies have emphasized the importance of executing smoother and more comfortable braking maneuvers to assist drivers in avoiding not only identified dangers but also collisions with subsequent vehicles \cite{bella2017effects}.

This work seeks to address these issues by formulating an optimal warning generation problem that considers the relation between the generated warning and the driver's reaction and also the interaction between the ego vehicle and other agents on a long horizon. The problem is modeled as a partially observed Markov decision process (POMDP), in which we quantify the value of warnings through the comfort and safety of the future ego vehicle trajectory in the context of surrounding objects, and the cost through their format and frequency. An optimal warning generation framework is proposed as a solution to the POMDP. The key contributions of this work are as follows:
\begin{itemize}[leftmargin=*]
    \item[-] We propose a novel formulation of the optimal warning generation problem that considers the driver and surrounding vehicle reactions, and both the safety and the comfort of future ego trajectories.
    \item[-] We propose a warning generation framework combining driver behavior estimation as the solution to the above problem. The framework has the flexibility to incorporate any prediction models of the driving scenario.
    \item[-] The proposed method is evaluated over comprehensive closed-loop simulation experiments, which demonstrates the superiority of the proposed solution to the existing warning generation methods. 
\end{itemize}

\section{RELATED WORKS}
\subsection{Driver warning system}
Most methods in the literature generate the warning through threshold-checking based on the current state. \cite{iranmanesh2018adaptive} generates the warnings by continuously comparing Time Headway with a flexible threshold that is updated based on drivers' actions. \cite{YUAN202064} generates warnings based on the distance identified from the front camera and the drivers' attention identified from the inside camera. Some methods consider a longer horizon by utilizing a learned neural network. The method in \cite{guo2022forwarding} and \cite{lee2018development} utilizes a neural network to predict the driver's behavior and generate warnings based on the TTC threshold. There are also some methods trying to solve the problem through vehicle-to-vehicle connections. In \cite{yang2020forward}, warnings are produced based on the driving intention of other vehicles which are transmitted to the ego vehicle through vehicle-to-vehicle (V2V) communication. 
\subsection{Human driving behavior with warnings}
There are studies in the literature that investigate the effect of the warning system on human driving behavior. In \cite{wu2021auditory}, researchers find that the human drivers' reactions to warnings vary with the type of warning. Meanwhile, the results in \cite{zhao2023effects} also indicate that age, year of driving experience, collision type, and warning type can affect driving performance. \cite{bella2017effects} shows that a good warning will produce a comfortable braking maneuver which not only helps to avoid forward collisions but also rear collisions.

\section{PROBLEM FORMULATION}\label{sec:formulation}

\begin{figure}[t]
\centerline{\includegraphics[width=1\linewidth]{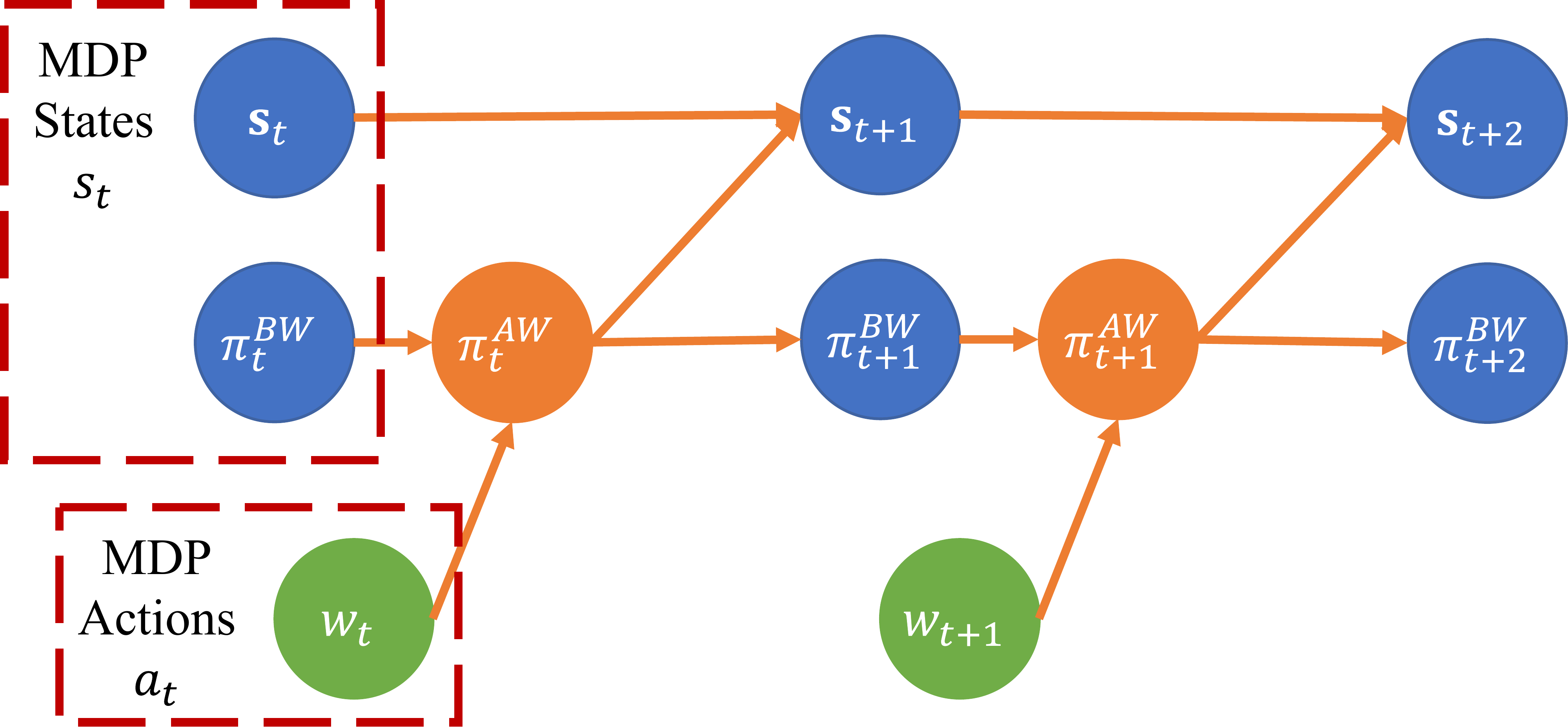}}
\caption{Markov decision process (MDP) formulation of the warning generation problem. The state of the MDP $s_t$, shown with blue, includes the state of the scenario $\mathbf{s}_t$ and ego driving policy before the warning $\pi^\text{BW}_t$. The action of the MDP $a_t$, shown with green, is the warning $w_t$ provided to the ego diver. The state transition, shown with orange, follows the Eq.~\ref{eq:StateTransition}.}
\label{fig:MDP_Formulation}
\end{figure}

\subsection{Ego Driving Behavior}
To provide necessary and accurate warning information to the ego driver, it is essential to model the ego driving behavior, which can be represented by a driving policy $\pi(\mathbf{a}|\mathbf{s}_t)$. $\mathbf{a}$ denotes the driver’s action including the acceleration and steering angle. $\mathbf{s}_t$ is the current state of the scenario, including ego dynamic states and its history $x_t$, all the $N$ surrounding agents’ dynamic states and their history $Y_t=\{y_t^1 , ... , y_t^N \}$, and all the other environment information like map and so forth.

\begin{table}
  \caption{Ego Driving Behaviors}
  \label{tab:EgoBehaviors}
  \centering
  \setlength{\tabcolsep}{10pt}
  \vspace{-6pt}
  \begin{tabular}{c|l}
    \toprule[1.0pt]
    Driving Behaviors & Description\\
    \midrule
    $\pi_\text{safe}$ & Driving with safe and optimal actions \\
    $\pi_\text{blind}$ & Driving without considering other agents \\
    $\pi_\text{brake}$ & Braking and then recovering to $\pi_\text{safe}$ \\
    $\pi_\text{delay}$ & Switching to new behaviors with a delay\\
  \bottomrule[1.2pt]
\end{tabular}
\vspace{-12pt}
\end{table}

We model the ego driving behavior into four different modes, as shown in Table~\ref{tab:EgoBehaviors}. Before the warning is provided, the ego driver may already notice the danger and drive safely considering all the other agents. The ego driving behavior under this mode is denoted by $\pi_\text{safe}(\mathbf{a}|\mathbf{s}_t= (x_t, Y_t))$. 

Meanwhile, the ego driver can also be distracted and carelessly ignore the danger, under which the policy can be denoted as $\pi_\text{blind}(\mathbf{a}|\mathbf{s}_t)$. Even though the driver is driving unsafely under $\pi_\text{blind}$, the driving behavior should still follow the human driving properties. Thus, we can derive $\pi_\text{blind}$ from $\pi_\text{safe}$ by masking out $Y_t$:
\begin{equation}
\pi_\text{blind}(\mathbf{a}|\mathbf{s}_t) = \pi_\text{safe}(\mathbf{a}|\mathbf{s}_t = (x_t, \emptyset)).
\end{equation}

After the warning, it can be observed that sometimes the driver tends to decelerate immediately before really considering the scene and optimizing their actions \cite{wu2021auditory}. The action policy on this mode is denoted by $\pi_{\text{brake}}(T_R)(\mathbf{a}|\mathbf{s}_t)$. Under this policy, the ego driver will take brake action for $T_R$ time, and recover to $\pi_\text{safe}$ after that:
\begin{equation}
\pi_{\text{brake}}(T_R)(\mathbf{a}|\mathbf{s}_t) = \begin{cases}
      \mathbf{a}_\text{decelerate}, & t < T_R\\
      \pi_\text{safe}(\mathbf{a}|\mathbf{s}_t = (x_t, Y_t)), & t \geq T_R \\
    \end{cases}.
\end{equation}
where $ \mathbf{a}_\text{decelerate}$ is a fixed action that has negative acceleration and $T_R$ is a fixed parameter as it is a feature of an individual. Note that $\pi_{\text{brake}}$ is always a sub-optimal compared with $\pi_\text{safe}$ as its action space during $T_R$ is a subset of $\pi_\text{safe}$. 

In addition to that, it also takes some time before the human really begins to react to the warning \cite{zhu2020impact,yue2021effects,bella2017effects}. To describe this behavior, we define the delay policy $\pi_\text{delay}(\pi_b,\pi_a,T_D)(\mathbf{a}|\mathbf{s}_t)$. When driving with this policy, the ego driver will follow $\pi_b$ for $T_D$ time and switch to $\pi_a$ after that, where $\pi_b$ and $\pi_a$ can be any driving policies except $\pi_\text{delay}$ itself:
\begin{equation}
\pi_\text{delay}(\pi_b,\pi_a,T_D)(\mathbf{a}|\mathbf{s}_t) = \begin{cases}
      \pi_b(\mathbf{a}|\mathbf{s}_t), & t < T_D\\
      \pi_a(\mathbf{a}|\mathbf{s}_t), & t \geq T_D \\
    \end{cases}.
\end{equation}
Similarly, we are also considering $T_D$ is fixed over different combinations of behaviors, as it is a feature of an individual.

In practice, those behaviors can be obtained through different methods, such as data-driven methods in \cite{suo2021trafficsim, varadarajan2022multipath++, chang2022analyzing,li2024residual} and model-based methods in \cite{zhan2017spatially, li2022efficient, schwarting2019social,doi:10.1177/02783649231215371}. We don't assume any specific methods in our framework.

\subsection{Behavior Transition}

After providing the warning information, there is a chance that the driver doesn't notice the warning and doesn't react to the warning \cite{yue2021effects, zhao2023effects, wu2021auditory}. Meanwhile, the different levels of warning can make drivers react to the scene in different ways \cite{wu2021auditory,bella2017effects}. Thus, it is crucial to model the behavior transition of the driver when receiving the warnings. Let
\begin{equation}
\mathcal{T}(\pi_t,w_t) = \text{Pr}(\pi|\pi_t,w_t)
\end{equation}
represent the probability of the driving policy $\pi_t$ switching to another policy $\pi$ after receiving the warning $w_t\in W$, where 
$W = \{\text{No warning}, \text{Text}, \text{Voice}, \text{Alarm}, \text{Take over}\}$ is the set of possible warnings that can be provided to the driver.

In practice, this model can be obtained through the human drivers' data. However, there is still some domain knowledge that needs to be included. When the scenario is extremely dangerous and the system decides to take over the control to force the vehicle to slow down, the driving policy will switch to $\pi_{\text{brake}}$ immediately without any delay. Meanwhile, as $\pi_\text{brake}$ is a more cautious behavior than $\pi_\text{safe}$, any warning can only transform $\pi_\text{safe}$ to $\pi_\text{brake}$ but not the opposite direction.

\subsection{System Modeling}
With the modeling of the ego driving behavior and its transition, the warning generation problem can be discretized and modeled as a Markov decision process (MDP) problem, as shown in Fig.~\ref{fig:MDP_Formulation}. The state of the MDP $s_t$ includes the state of the scenario $\mathbf{s}_t$ and ego driving policy before the warning $\pi^\text{BW}_t$; the action of the MDP $a_t$ is the warning $w_t$ provided to the ego diver: 
\begin{equation}\label{eq:MDPState}
\begin{aligned}
    s_t &= (\mathbf{s}_t,\pi^\text{BW}_{t}),\\
    a_t &= w_t.
\end{aligned}
\end{equation}
During the MDP state transition, the ego driving policy after the warning $\pi^\text{AW}_t$ will be decided first following the behavior transition model $\mathcal{T}(\pi^\text{BW}_t,w_t)$. Then, the policy will produce the driver's action and move the ego dynamic state to the next step through the dynamic. Meanwhile, the surrounding agents will also decide their actions based on the current state of the scenario. As the step time between the two states is small enough, we assume the driving policy at the next step before the warning is the same as the driving policy at the current step after the warning. Thus, the whole state transition process of the proposed MDP can be summarized as:
\begin{equation}\label{eq:StateTransition}
\begin{aligned}
    \pi^\text{AW}_t &\sim \mathcal{T}(\pi^\text{BW}_t,w_t), \\
    \mathbf{a}_t & \sim \pi^\text{AW}_t(\mathbf{a}|\mathbf{s}_t), \\
    x_{t+1} &= f(x_{t},\mathbf{a}_t), \\
    Y_{t+1} &= g(\mathbf{s}_t), \\
    \pi^\text{BW}_{t+1} &= \pi^\text{AW}_t,
\end{aligned}
\end{equation}
where $g$ represents the behavior and dynamic of surrounding agents, which can be obtained through different existing trajectory prediction methods in the literature \cite{salzmann2020trajectron++, gu2021densetnt, chang2023editing}.

The reward of the proposed MDP $R(s_t,w_t)$ contain two aspects: the trajectory reward $R_\text{Traj}(\mathbf{s}_t,\mathbf{a}_t)$ and the cost of warning action $R_\text{warning}(w_t)$. The trajectory reward quantifies the safety and comfort of the trajectory, the efficiency of the driving action, and the tendency to follow the desired velocity, while the cost of warning action quantifies the severeness of a warning. For example, an alarm warning is way more severe and discomforts the driver more than a text message warning. Let $H$ denote the searching horizon, the optimization problem we are solving can then be written as:
\begin{equation}
\begin{aligned}
    \max_{w_0,...,w_H} & \mathbb{E}\left[ \sum_{t=0}^H \gamma^{t} R(s_t,w_t)\right], \\
    s.t. & \  \ (\ref{eq:StateTransition})
\end{aligned}
\end{equation}
where $s_t = (\mathbf{s}_t, \pi^\text{BW}_t)$ is the state of MDP as defined above, and $\gamma$ is a discount factor. However, in general, the current ego driving behavior $\pi_t^\text{BW}$ is unknown, which makes the problem become a partially observed MDP (POMDP) and be very hard to solve. In Sec.~\ref{sec:BehaviorEstimation}, \ref{sec:WarningSearcher}, we will discuss how we handle this problem in detail.

\section{BEHAVIOR ESTIMATION} \label{sec:BehaviorEstimation}
As discussed in the previous section, in practice, it can be difficult for the system to directly obtain the current ego driving behavior. However, the states of the scenario $\mathbf{s}_t$ as well as the driver's actions $\mathbf{a}_t$ are easy to access through sensors. Therefore, we can estimate the ego driver's behavior through the state measurements and driver's actions by Bayesian inference. The estimation is computed with three steps: model prediction, observation correction, and state transition. These steps compose a generic design, thus when we have some other observations of the driving behavior like head pose and and gaze \cite{iranmanesh2018adaptive, wu2022toward}, it can be fused into these steps easily. 
\subsubsection{Model prediction}
Let $b(\pi_t^\text{BW})$ denote the estimated behavior distribution, and $b(\pi_0^\text{BW})$ be the initial estimated distribution. After receiving the warning, the estimation will be updated with the model:
\begin{equation}\label{eq:PredictionUpdate}
\begin{aligned}
    b^-(\pi_t^\text{AW}) = \sum_{\pi \in \Pi} \mathcal{T}(\pi,w_t) b(\pi_t^\text{BW} = \pi),
\end{aligned}
\end{equation}
where $b^-(\pi_t^\text{AW})$ is the estimated distribution after the model prediction step, $\Pi$ is the set that contains all possible driving policies.
\subsubsection{Observation correction}
After the ego driver takes an action $\mathbf{a}_t$, the estimated behavior will be updated through the Bayesian inference:
\begin{equation}\label{eq:CorrectionUpdate}
\begin{aligned}
    b^+(\pi_t^\text{AW} = \pi) = {{\pi(\mathbf{a}_t|\mathbf{s}_t) \times b^-(\pi_t^\text{AW}=\pi)} \over {\sum_{\pi \in \Pi}} \pi(\mathbf{a}_t|\mathbf{s}_t) \times b^-(\pi_t^\text{AW}=\pi) },
\end{aligned}
\end{equation}
where $b^+(\pi_t^\text{AW})$ is the estimated distribution after the observation correction step.
\subsubsection{State transition}
Based on the state transition modeling Eq.~(\ref{eq:StateTransition}), the estimated distribution of the behavior at the next step before the warning can be obtained by:
\begin{equation}\label{eq:StateTransitionUpdate}
\begin{aligned}
    b(\pi_{t+1}^\text{BW}) =  b^+(\pi_t^\text{AW}).
\end{aligned}
\end{equation}
Note that for behavior $\pi_\text{brake}$ and $\pi_\text{delay}$ that have internal behavior transitions, such transitions may happen as time passes and will be captured during this state transition update step. 

\section{OPTIMAL WARNING SEARCHER} \label{sec:WarningSearcher}
With the estimated behavior distribution, the problem can be solved either through the MDP formulation which utilizes the most probable behavior from estimation, or POMDP which considers the whole estimated distribution and solves the optimization over the belief space. However, due to the complexity of the problem and the exponential tendency of the development of future states, it is infeasible to solve the exact POMDP problem in real-time. In the following subsections, we investigate two different approximated solutions to the POMDP problem.

\begin{figure}[t]
\centerline{\includegraphics[width=0.85\linewidth]{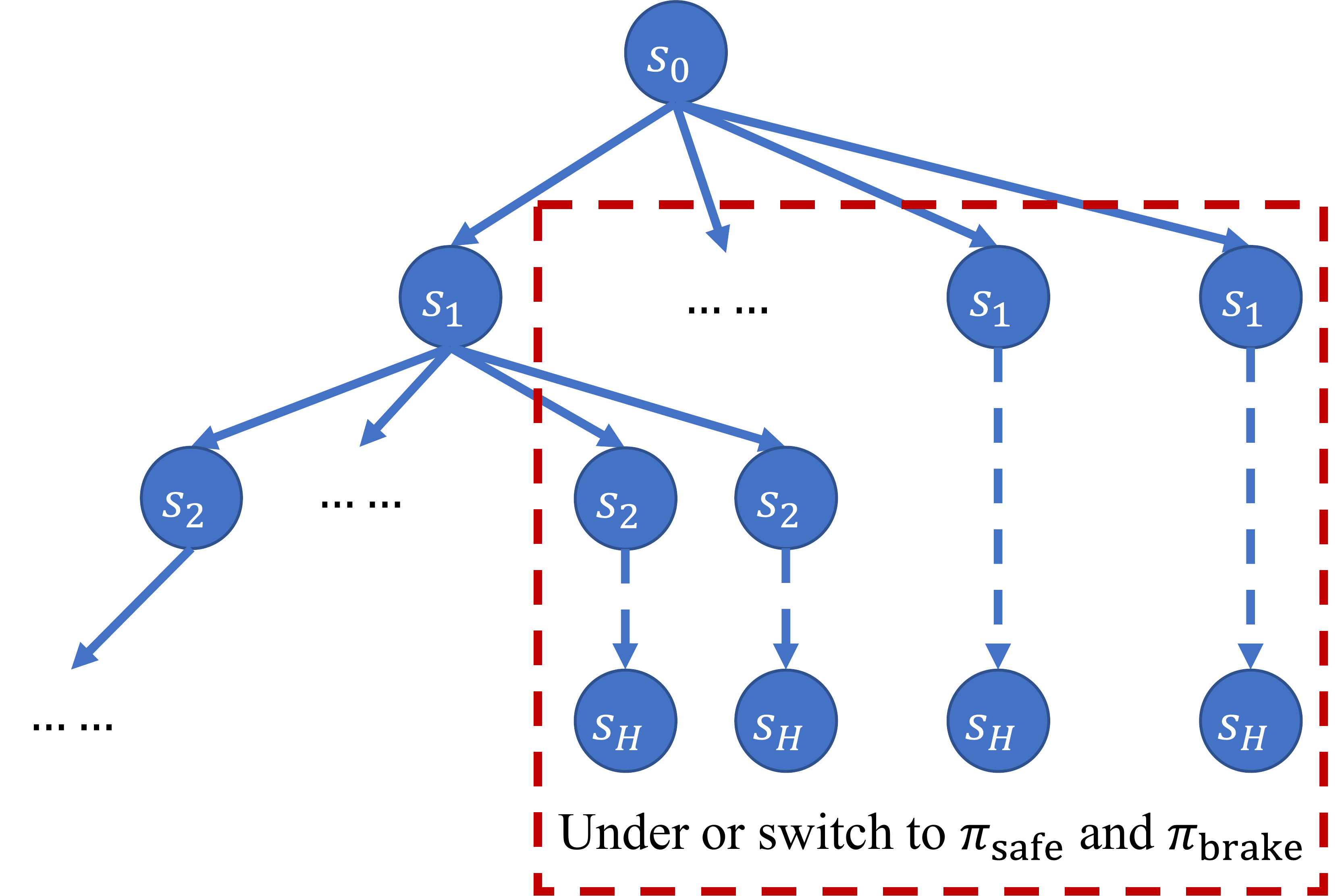}}
\caption{MDP State Tree After Simplification Rule Applied. Each node represents an MDP state. Each edge represents a possible state transition.}
\label{fig:ReducedTree}
\end{figure}

\begin{algorithm}[t]
    \caption{Optimal Warning Searcher}
    \label{alg:WarningSearcher}
    \textbf{Input:}
    Current state of the scenario $\mathbf{s}_0$, estimate of current ego driving policy $\hat{\pi}^\text{BW}_0$, behavior transition model $\mathcal{T}(\pi^\text{BW}_t,w_t)$, searching horizon $H$.\newline
    \textbf{Output:}
    Root state of built tree $s_0$, warning sequence for future steps $\{w_0,...,w_{H-1}\}$. 
    \begin{algorithmic}[1]
    \State Initialize the root of the MDP state tree: $s_0 = (\mathbf{s}_0,\pi^\text{BW}_0)$
    \State $s_H\leftarrow \Call{ForwardSimulation}{s_0}$
    \State $\{w_0,...,w_{H-1}\} \leftarrow \Call{BackPropagation}{s_H}$
    \State \Return $s_0, \{w_0,...,w_{H-1}\}$
    \end{algorithmic}
\end{algorithm}

\subsection{MDP with estimated state}
After acquiring the estimated behavior distribution, we can extract an estimate from the distribution as an estimated state and solve the problem as MDP. The behavior estimates are generated by the following equation:
\begin{equation}\label{eq:BehaviorEstimates}
\begin{aligned}
    \hat{\pi}^\text{BW}_t =  \begin{cases}
      \pi_\text{blind}, \ \ b({\pi}^\text{BW}_t=\pi_\text{blind}) > \text{Th}_\text{safety}\\
      \arg\max_{\pi} b({\pi}^\text{BW}_t=\pi), \ \ otherwsie\\
    \end{cases},
\end{aligned}
\end{equation}
where $\hat{\pi}^\text{BW}_t$ is the extracted estimates at time step $t$, $\text{Th}_\text{safety}$ is a safety threshold to satisfy the robustness and safety requirement of the warning system.

By considering $\hat{\pi}^\text{BW}_t$ as an estimated state, we can solve the problem by estimating the Bellman equation online. Based on the modeling presented in Sec.~\ref{sec:formulation}, the Q-function can be represented as follows:
\begin{equation}\label{eq:BellmanEquation}
\begin{aligned}
    Q(s_t,a_t)  & = Q\left((\mathbf{s}_t,\pi^\text{BW}_t),w_t\right) \\
                & = R_\text{warning}(w_t) + \mathbb{E}_{(*)}\left[R_\text{Traj}(\mathbf{s}_t,\mathbf{a}_t)\right]\\
                & + \gamma \mathbb{E}_{(\ref{eq:StateTransition})}\left[ V^*((\mathbf{s}_{t+1},\pi^\text{BW}_{t+1}))\right], \\
    V^*(s_t)      & = V^*((\mathbf{s}_{t},\pi^\text{BW}_{t})) \\
                & = \max_{w_{t}}Q\left((\mathbf{s}_{t},\pi^\text{BW}_{t}),w_{t} \right),
\end{aligned}
\end{equation}
where,
$$(*): \pi^\text{AW}_t \sim \mathcal{T}(\pi^\text{BW}_t,w_t), \ \mathbf{a}_t\sim \pi^\text{AW}_t(\mathbf{a}|\mathbf{s}_t)$$

By discretization of actions, the development of the states can be formed as a tree, as shown in Fig.~\ref{fig:ReducedTree} in which each node represents an MDP state, and each edge represents a transition between the states. Based on the tree, we proposed a search algorithm to estimate the Q-value of each action at each state. The proposed algorithm \ref{alg:WarningSearcher} contains two phases: forward simulation and back propagation. 

\begin{algorithm}[t]
    \caption{Optimal Warning Searcher: Forward Simulation}
    \label{alg:Function:Forward}    
    \begin{algorithmic}[1]
    \Function{ForwardSimulation}{$s_0$}
    \State $t \leftarrow 0$
    \While{$t<H$}
        \State $s_t \leftarrow $ child of $s_{t-1}$ that $\pi^\text{BW}_{t} = \pi_\text{blind}$
        \State Initialize behavior list $\Pi^\text{AW}_t$ from $\mathcal{T}(\pi^\text{BW}_t,w_t)$
        \For{$\pi \in \Pi^\text{AW}_t$}
            \State $\mathbf{a} \leftarrow       \arg\max_{\mathbf{a}}\pi(\mathbf{a}|\mathbf{s}_t)$
            \State $x_{t+1} = f(x_{t},\mathbf{a})$
            \State $Y_{t+1} = g(\mathbf{s}_t)$
            \State $s_{t+1} \leftarrow (\mathbf{s}_{t+1} = (x_{t+1}, Y_{t+1}),\pi^\text{BW}_{t+1} = \pi)$
            \State $r(s_t,s_{t+1}) \leftarrow R_\text{Traj}(\mathbf{s}_t,\mathbf{a})$
            \State Add $s_{t+1}$ to the children list of $s_t$
            \If{$\pi$ is switching to $\pi_\text{safe}$ or $\pi_\text{brake}$}
                \State Build full branch form $s_{t+1}$ to $s_H$ with $\pi$
            \EndIf
        \EndFor
    \State $t \leftarrow t+1$
    \EndWhile
    \State \Return $s_H$ that $\pi^\text{BW}_{H} = \pi_\text{blind}$
    \EndFunction
    \end{algorithmic}
\end{algorithm}
\subsubsection{Forward simulation} During the forward simulation, the tree will be constructed from the current state with driving actions with maximum probability from the different driving policies. Even though the driving actions are simplified, the states will still develop exponentially and become intractable after several time steps. To reduce the dimension of the tree, we simplify the branches whose ego driving behavior is under or switching to $\pi_\text{safe}$ and $\pi_\text{brake}$. On those branches, the warning provided to the driver is fixed to $w_t = \text{No warning}$. Therefore, for those branches, the algorithm will roll out to the end and return the Q-value immediately, which reduces the treewidth over searching depth from exponential increasing to linear increasing, as shown in Fig.~\ref{fig:ReducedTree}. 

Such simplification is reasonable because of the following reasons. As introduced in Sec.~\ref{sec:formulation}, $\pi_\text{safe}$ and $\pi_\text{brake}$ are policies that will decide the driving actions while considering all the other agents. The $\pi_\text{brake}$ is always sub-optimal compared with $\pi_\text{safe}$ and can not be transformed to $\pi_\text{safe}$ through warning. Thus, as long as these behaviors are able to produce a collision-free trajectory after the delay, there is no need for a further warning. When the delay of reaction makes $\pi_\text{safe}$ and $\pi_\text{brake}$ fail to produce a feasible trajectory, only taking over the control of the vehicle can handle the scenario. Under this case, the taking over can be applied at an early step or late step based on the rewards. However, for safety considerations, we would apply taking over earlier so that the ego driver has more time and space to react. Thus, with current simplification, when we know the policies will fail due to the delay, we will apply taking over directly.

\begin{algorithm}[t]
    \caption{Optimal Warning Searcher: Back Propagation}
    \label{alg:Function:BackPropagation}    
    \begin{algorithmic}[1]
    \Function{{BackPropagation}}{$s_H$}
    \State $V(s_H) = 0$
    \State $s_t \leftarrow s_H$
    \While{$s_t$ has parent}
        \State $s_p \leftarrow $ parent of $s_t$ 
        \State Extract $r(s_p,s_{p+1}),\forall s_{p+1}\in$ children of $s_p$ 
        \For{$w_p \in W$}
            \State $Q\left((\mathbf{s}_p,\pi^\text{BW}_p),w_p\right) \leftarrow$ Update by Eq.~\ref{eq:BellmanEquation}
        \EndFor
        \State $w_p \leftarrow \arg\max_{w_{p}}Q\left((\mathbf{s}_{p},\pi^\text{BW}_{p}),w_{p} \right)$ 
        \State $V(s_p) \leftarrow \max_{w_{p}}Q\left((\mathbf{s}_{p},\pi^\text{BW}_{p}),w_{p} \right)$
        \State $s_t \leftarrow s_p$ 
    \EndWhile
    \State \Return $\{w_0,...,w_{H-1}\}$
    \EndFunction
    \end{algorithmic}
\end{algorithm}
\subsubsection{Back propagation} During the back propagation, the estimated Q-value will be computed backward from the leaf nodes. In the meantime, it will also select the best warning $w_t$ to be provided at each time step. After propagating to the root node, the algorithm has updated all Q-values and is able to return a warning sequence for future steps when the ego driver doesn't notice the warning.

\subsection{Approximated POMDP}
The estimated state ignores the uncertainty and has to utilize the safety threshold $\text{Th}_\text{safety}$ to reduce the error brought by the overconfident estimates. To address this problem, we can build another approximation method by solving the warning with the maximum expected Q-value over the estimated behavior distribution. The formulation can be represented as:
\begin{equation}
\begin{aligned}
    \max_{w_0} & E_{\pi^\text{BW} \sim b(\pi^\text{BW}_0)}\left[ Q\left((\mathbf{s}_0,\pi^\text{BW}),w_0\right) \right], 
\end{aligned}
\end{equation}
where each Q-value $Q\left((\mathbf{s}_0,\pi^\text{BW}),w_0 \right)$ can be obtained through the same algorithm for the MDP with estimated state.

\section{EXPERIMENTS}
In this section, we conduct several experiments to evaluate the proposed framework. In Sec.~\ref{sec:expsetting}, we introduce the environment setting and configurations of the proposed framework for experiments and the rule-based warning generator baseline. In Sec.~\ref{sec:expresult}, we present the closed-loop simulations and provide an analysis of the performance.

\begin{figure}[t]
\centerline{\includegraphics[width=0.9\linewidth]{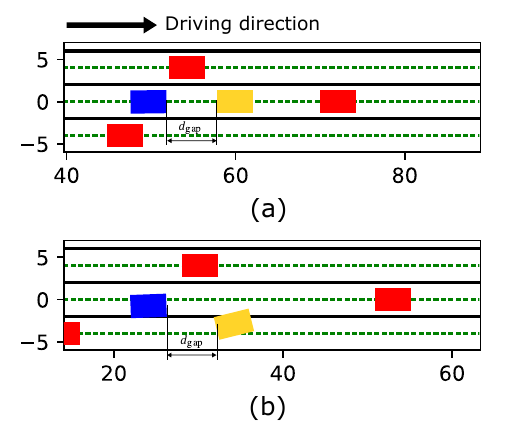}}
\caption{Experiment scenarios. The blue vehicle represents the ego vehicle driving at $11 m/s$. The red vehicles are background vehicles. The yellow vehicle is the active dangerous vehicle that will trigger the warning to the ego vehicle: (a) it will conduct a hard brake decreasing its speed from $12 m/s$ to $8m/s$ with maximum deceleration; (b) it will switch to the ego vehicle's lane at a lower speed $8 m/s$. $d_\text{gap}$ represents the gap between the ego vehicle and the active dangerous vehicle.}
\label{fig:environments}
\vspace{-12pt}
\end{figure}

\subsection{Experiment setting}\label{sec:expsetting}
\subsubsection{Scenarios}
We are primarily interested in those scenarios that are dangerous to the ego driver and focus our experiments on the lane change scenarios and hard braked front vehicle scenarios, as shown in Fig.~\ref{fig:environments}. During the simulation, the proposed framework is applied to the ego vehicle whose initial driving policy is $\pi_\text{blind}$. We also create an active dangerous vehicle with some background vehicles, which will cause danger and collisions if the ego driver doesn't change his behavior. Let $d_\text{gap}$ represent the gap between the ego vehicle and the active dangerous vehicle. The experiments are conducted over different initial $d_\text{gap}$, while the behavior is simulated by the IDM model with parameters fitting from real-world data.

\subsubsection{Baseline}
We compare our method to the classical Time to Collision (TTC) based warning, in which the warning is decided based on a TTC threshold. In addition to that, we also implement an adaptive rule-based warning generator baseline\cite{yang2020forward}. The baseline method utilizes the speed and gap between the vehicles to compute the minimum gap $d_\text{min}$, assuming a hard braking happened. Let $v_\text{front}$ and $v_\text{ego}$ denote the longitudinal speed of the front vehicle and ego vehicle respectively. Let $acc_\text{min}$ represent the minimum acceleration vehicles can achieve. Thus the computation can be summarized as:
\begin{equation}
\begin{aligned}
    d_\text{front} & = {{v_\text{front}^2}\over {2 |acc_\text{min}|}}, \\
    d_\text{ego} & = v_\text{ego} T_D + {{v_\text{ego}^2}\over {2 |acc_\text{min}|}},\\
    d_\text{min} & = d_\text{gap} + d_\text{front} - d_\text{ego},
\end{aligned}
\end{equation}
where $d_\text{front}$ and $d_\text{ego}$ represent the distance that the front vehicle and ego vehicle will travel when doing hard braking respectively. Then, the warning can be generated by evaluating the following inequality:
\begin{equation}
\begin{aligned}
    & d_\text{min} \leq - \alpha_w v_\text{ego} T_D, \\
\end{aligned}
\end{equation}
where $\alpha_w$ is a parameter varying with the severeness of a warning. When $\alpha_w = 1$, it represents the condition for the system to take over the control:
\begin{equation}
\begin{aligned}
    & d_\text{gap} + {{v_\text{front}^2}\over {2 |acc_\text{min}|}} - {{v_\text{ego}^2}\over {2 |acc_\text{min}|}} \leq 0, \\
\end{aligned}
\end{equation}
The left half of the equation represents the minimum gap when the system takes over the control to conduct a hard brake without delay. The condition prevents the ego vehicle from entering this dangerous zone.

\subsubsection{Reward design}
The proposed framework chooses the warning based on the trajectory reward $R_\text{Traj}$ and the cost of warning action $R_\text{warning}$. In experiments, $R_\text{Traj}$ is defined as:
\begin{equation}
\begin{aligned}
    & R_\text{Traj}(\mathbf{s}_t,\mathbf{a}_t) = -w_v (v_{t}-v_\text{desire})^2 - w_{acc} acc_t^2 - I(\mathbf{s}_{t}),
\end{aligned}
\end{equation}
where $v_{t}$ and $acc_t$ are the longitudinal velocity and acceleration of the vehicle respectively. $v_\text{desire}$ denotes the desired longitudinal velocity of the vehicle. $I(\mathbf{s}_{t})$ is an indicator function that will return infinity when a collision happens and zero otherwise. $w_v$ and $w_{acc}$ are parameters to balance the weight of velocity and acceleration. 

During experiments, we use $w_v = 0.5$, $w_{acc} = 0.1$, $v_\text{desire} = 11 m/s$. The cost of warning action $R_\text{warning}$ describes the severeness of a warning. During experiments, $R_\text{warning}(w_t)$ is defined as 
$\{\text{No warning}: 0, \text{Text}: -1, \text{Voice}: -20, \text{Alarm}: -50, \text{Take over}: -10^8\}$. The cost of taking over the control is set to large so that it will be only applied when other actions are infeasible.

\begin{table}[t]
\centering
\caption{Trajectory Reward in Closed-loop Experiments} \label{tab:WarningRewards}
\begin{tabular}{@{}clcc@{}}
\toprule
\multirow{2}{*}{$d_\text{gap}(0)$} & \multicolumn{1}{c}{\multirow{2}{*}{Method}} & \multicolumn{2}{c}{{Average Trajectory Reward $\bar R_\text{Traj}$}} \\ \cmidrule(l){3-4}
 & & Front Hard Brake  & Lane Change \\ \midrule
\multirow{3}{*}{$8.5m$} & Estimated State MDP & $-1038.06\pm 7.08$ & $-1184.02\pm 0.28$  \\
                          & Appr. POMDP  & ${-1040.53 \pm 17.14}$ & $-1183.92 \pm 0.19$ \\ 
                          & TTC Baseline  & ${-1307.50 \pm 49.69}$ & $-1411.25 \pm 0.00$ \\
                          & Rule-based Baseline  & ${-1236.61 \pm 109.50}$ & $-1411.25 \pm 0.00$  \\  \midrule
                          
\multirow{3}{*}{$13.5m$} & Estimated State MDP & $-762.45\pm 35.54$ & $-806.56\pm 35.88$ \\
                          & Appr. POMDP & ${-767.80 \pm 46.17}$ & $-857.77 \pm 50.23$  \\
                          & TTC Baseline  & ${-1073.82 \pm 38.44}$ & $-1156.52 \pm 64.95$ \\
                          & Rule-based Baseline  & ${-883.48 \pm 83.61}$ & $-948.64 \pm 19.90$  \\  \midrule
                          
\multirow{3}{*}{$18.5m$} & Estimated State MDP & $-632.48\pm 43.97$ & $-678.56\pm 30.18$ \\
                          & Appr. POMDP & ${-628.06 \pm 54.68}$ & $-678.29 \pm 40.41$  \\ 
                          & TTC Baseline  & ${-757.15 \pm 39.40}$ & $-873.34 \pm 71.90$ \\
                          & Rule-based Baseline  & ${-648.32 \pm 76.20}$ & $-697.07 \pm 72.60$ \\ \bottomrule
\end{tabular}
\end{table}

\begin{table}[t]
\centering
\caption{Closed-loop Warning Experimental Results of Lane Change} \label{tab:WarningCounts}
\begin{tabular}{@{}clcccc@{}}
\toprule
\multirow{2}{*}{$d_\text{gap}(0)$} & \multicolumn{1}{c}{\multirow{2}{*}{Method}}   & \multicolumn{4}{c}{Average Count of Warning} \\ \cmidrule(l){3-6}
 & &  Text & Voice & Alarm & Take Over \\ \midrule
\multirow{3}{*}{$8.5m$} & Estimated State MDP  &$0.00$ & $1.00$ & $1.00$ & $0.00$  \\
                          & Appr. POMDP & $0.00$ & $1.00$ & $1.07$ & $0.00$ \\ 
                          & TTC Baseline & $0.00$ & $2.00$ & $1.00$ & $1.00$  \\
                          & Rule-based Baseline & $0.00$ & $0.00$ & $4.00$ & $0.00$  \\  \midrule
                          
\multirow{3}{*}{$13.5m$} & Estimated State MDP  & $1.00$ & $2.12$ & $0.01$ & $0.00$  \\
                          & Appr. POMDP  & $2.71$ & $1.00$ & $0.51$ & $0.00$\\
                          & TTC Baseline & $1.67$ & $0.93$ & $0.94$ & $0.67$  \\
                          & Rule-based Baseline  & $0.96$ & $2.04$ & $1.04$ & $0.00$ \\  \midrule
                          
\multirow{3}{*}{$18.5m$} & Estimated State MDP &$0.00$ & $2.10$ & $0.01$ & $0.00$\\
                          & Appr. POMDP & $0.99$ & $1.12$ & $0.13$  & $0.00$ \\
                          & TTC Baseline & $0.94$ & $0.94$ & $0.94$ & $0.52$  \\
                          & Rule-based Baseline & $3.66$ & $1.47$ & $0.01$  & $0.00$  \\ \bottomrule
\end{tabular}
\vspace{-12pt}
\end{table}

\subsection{Closed loop analysis}\label{sec:expresult}
To focus our study on the proposed framework, during the simulation, the prediction model is assumed to be effective enough to capture the accurate trajectories of surrounding agents. The result of each scenario is computed over 8-second long trajectories and averaged over 200 simulations. The step horizon of the warning generation $H$ is $10$ while the duration of each time step is $0.5$ second. Thus the system will simulate $5$ seconds into the future to provide the warning. During the experiment, we run the warning search every $0.5$ second, in a model predictive control manner.

\subsubsection{Behavior estimation}
\begin{figure}[t]
\centerline{\includegraphics[width=1.0\linewidth]{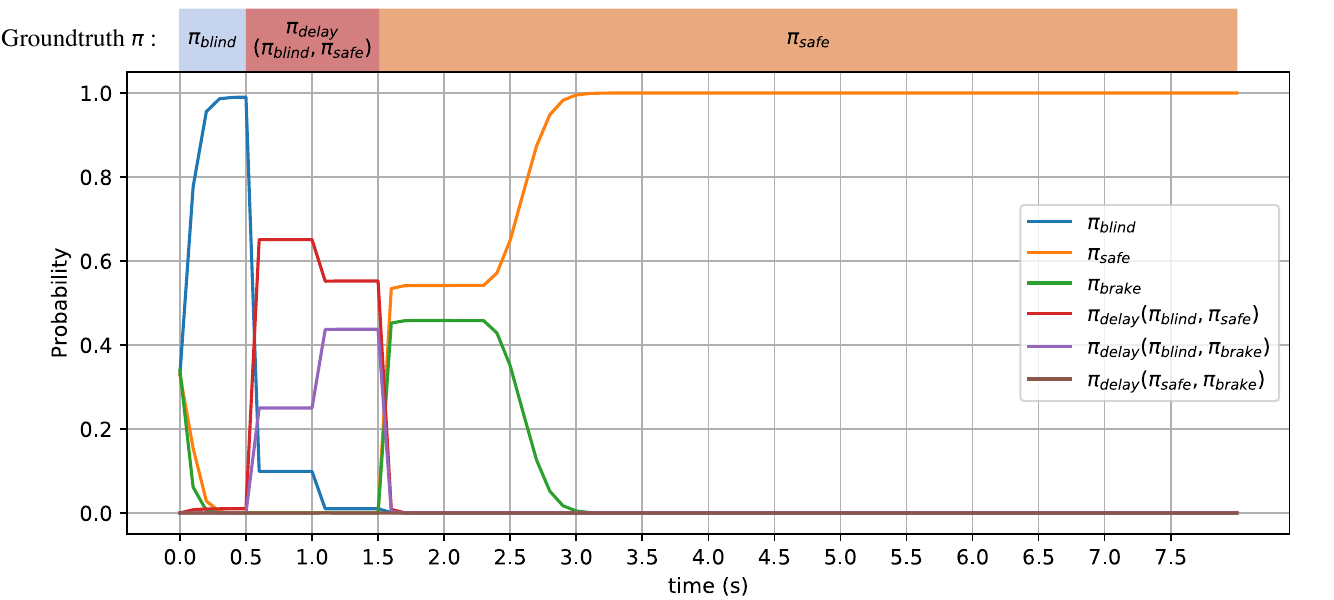}}
\caption{Result of proposed behavior estimation. The ground truth of current behavior is shown at the top of the plot. Two voice warnings are provided at $0.5s$ and $1s$.}
\label{fig:BehavorEstimation}
\vspace{-12pt}
\end{figure}

To evaluate the proposed behavior estimation, we plot the estimated distribution at each time step. As shown in Fig.~\ref{fig:BehavorEstimation}, the proposed behavior estimation is able to capture the correct current behavior. When the vehicle switches its behavior with a delay, there is less information in the observed action as the actual driving policy hasn't changed. Therefore, the estimation relies on the model prediction step as shown with the curve during $0$ to $1.5$ second. After the delay time, the policy can be identified through the action with the observation correction step since the $\pi_\text{blind}$ will not react to the danger as shown with the curve during $1.5$ to $3.0$ second.

\subsubsection{Trajectory performance}
\begin{figure}[t]
\centerline{\includegraphics[width=1.0\linewidth]{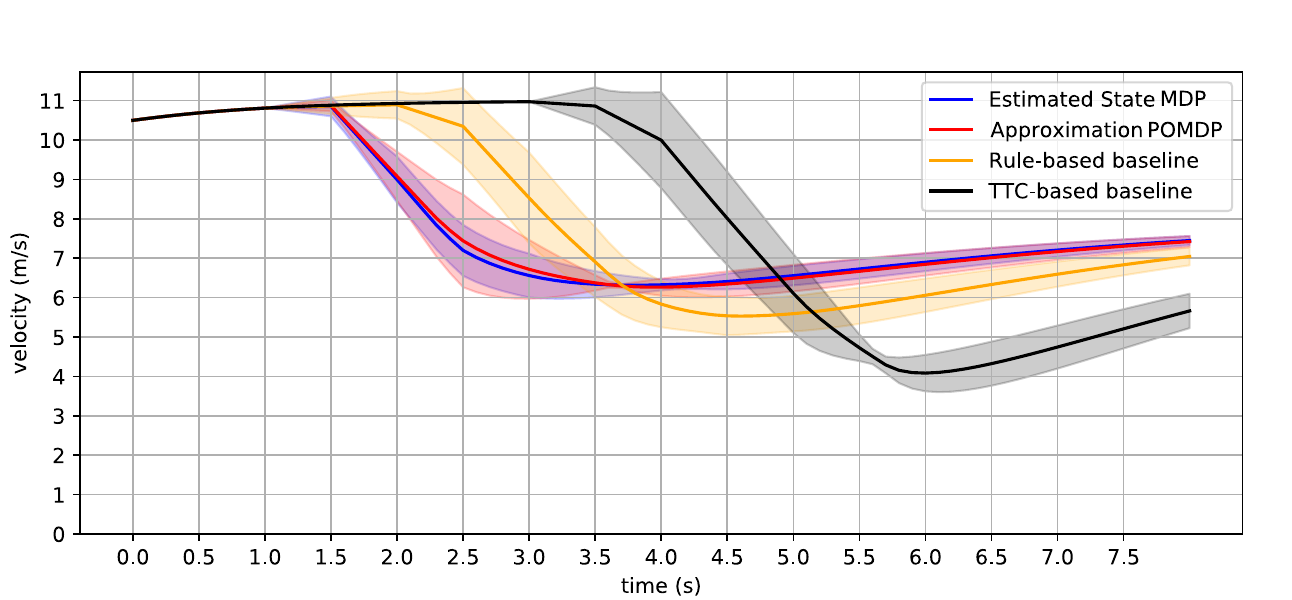}}
\caption{Closed-loop speed profile of front hard brake scenario with initial $d_\text{gap} = 13.5m$.}
\label{fig:BehavorSpeed}
\vspace{-12pt}
\end{figure}

To evaluate the proposed warning generation framework, we compared the reward of the closed-loop trajectories in both scenarios. As shown in Table~\ref{tab:WarningRewards}, the proposed framework can consistently achieve higher trajectory rewards than baselines in both scenarios over different initial $d_\text{gap}$. 
Meanwhile, as it can be seen in Fig.~\ref{fig:BehavorSpeed}, the proposed method provides warnings earlier and leads to smoother trajectories. 

\subsubsection{Warning efficiency}
We also evaluate the warning efficiency by the count of warnings over different $d_\text{gap}$ during the experiments. As can be seen in Table~\ref{tab:WarningCounts}, since the rule-based baseline doesn't model the ego driver's behavior and transition, it tends to give more warning than the proposed framework. The proposed method shows better warning efficiency over the baseline.

\section{CONCLUSIONS}
In this work, we study the problem of optimally generating driver warnings by considering the interactions between the generated warning and the driver behavior. We propose a flexible POMDP formulation with an optimal warning generation framework to solve the proposed problem. Simulation results demonstrate the superiority of the proposed solution to the existing warning generation methods. 

\addtolength{\textheight}{-8cm}   


\bibliographystyle{IEEEtran}
\bibliography{ref}{}

\begin{thebibliography}{10}
\providecommand{\url}[1]{#1}
\csname url@samestyle\endcsname
\providecommand{\newblock}{\relax}
\providecommand{\bibinfo}[2]{#2}
\providecommand{\BIBentrySTDinterwordspacing}{\spaceskip=0pt\relax}
\providecommand{\BIBentryALTinterwordstretchfactor}{4}
\providecommand{\BIBentryALTinterwordspacing}{\spaceskip=\fontdimen2\font plus
\BIBentryALTinterwordstretchfactor\fontdimen3\font minus
  \fontdimen4\font\relax}
\providecommand{\BIBforeignlanguage}[2]{{%
\expandafter\ifx\csname l@#1\endcsname\relax
\typeout{** WARNING: IEEEtran.bst: No hyphenation pattern has been}%
\typeout{** loaded for the language `#1'. Using the pattern for}%
\typeout{** the default language instead.}%
\else
\language=\csname l@#1\endcsname
\fi
#2}}
\providecommand{\BIBdecl}{\relax}
\BIBdecl

\bibitem{zhu2020impact}
M.~Zhu, X.~Wang, and J.~Hu, ``Impact on car following behavior of a forward
  collision warning system with headway monitoring,'' \emph{Transportation
  research part C: emerging technologies}, vol. 111, pp. 226--244, 2020.

\bibitem{yue2021effects}
L.~Yue, M.~Abdel-Aty, Y.~Wu, J.~Ugan, and C.~Yuan, ``Effects of forward
  collision warning technology in different pre-crash scenarios,''
  \emph{Transportation research part F: traffic psychology and behaviour},
  vol.~76, pp. 336--352, 2021.

\bibitem{kamrani2023drivers}
M.~Kamrani, S.~Concas, A.~Kourtellis, M.~Rabbani, V.~C. Kummetha, and O.~Dokur,
  ``Drivers’ reactions to connected vehicle forward collision warnings:
  Leveraging real-world data from the thea cv pilot,'' \emph{Transportation
  research part F: traffic psychology and behaviour}, vol.~92, pp. 108--120,
  2023.

\bibitem{wu2021auditory}
X.~Wu and L.~N. Boyle, ``Auditory messages for intersection movement assist
  (ima) systems: effects of speech-and nonspeech-based cues,'' \emph{Human
  factors}, vol.~63, no.~2, pp. 336--347, 2021.

\bibitem{zhao2023effects}
W.~Zhao, S.~Gong, D.~Zhao, F.~Liu, N.~Sze, and H.~Huang, ``Effects of collision
  warning characteristics on driving behaviors and safety in connected vehicle
  environments,'' \emph{Accident Analysis \& Prevention}, vol. 186, p. 107053,
  2023.

\bibitem{TSMC96}
P.~An and C.~Harris, ``An intelligent driver warning system for vehicle
  collision avoidance,'' \emph{IEEE Transactions on Systems, Man, and
  Cybernetics - Part A: Systems and Humans}, vol.~26, no.~2, pp. 254--261,
  1996.

\bibitem{iranmanesh2018adaptive}
S.~M. Iranmanesh, H.~N. Mahjoub, H.~Kazemi, and Y.~P. Fallah, ``An adaptive
  forward collision warning framework design based on driver distraction,''
  \emph{IEEE Transactions on Intelligent Transportation Systems}, vol.~19,
  no.~12, pp. 3925--3934, 2018.

\bibitem{YUAN202064}
Y.~Yuan, Y.~Lu, and Q.~Wang, ``Adaptive forward vehicle collision warning based
  on driving behavior,'' \emph{Neurocomputing}, vol. 408, pp. 64--71, 2020.

\bibitem{pyo2016front}
J.~Pyo, J.~Bang, and Y.~Jeong, ``Front collision warning based on vehicle
  detection using cnn,'' in \emph{2016 International SoC Design Conference
  (ISOCC)}.\hskip 1em plus 0.5em minus 0.4em\relax IEEE, 2016, pp. 163--164.

\bibitem{yang2020forward}
W.~Yang, B.~Wan, and X.~Qu, ``A forward collision warning system using driving
  intention recognition of the front vehicle and v2v communication,''
  \emph{IEEE Access}, vol.~8, pp. 11\,268--11\,278, 2020.

\bibitem{guo2022forwarding}
L.~Guo, Y.~Jia, X.~Hu, and F.~Dong, ``Forwarding collision assessment with the
  localization information using the machine learning method,'' \emph{Journal
  of advanced transportation}, vol. 2022, 2022.

\bibitem{bella2017effects}
F.~Bella and M.~Silvestri, ``Effects of directional auditory and visual
  warnings at intersections on reaction times and speed reduction times,''
  \emph{Transportation research part F: traffic psychology and behaviour},
  vol.~51, pp. 88--102, 2017.

\bibitem{lee2018development}
S.~H. Lee, S.~Lee, and M.~H. Kim, ``Development of a driving behavior-based
  collision warning system using a neural network,'' \emph{International
  journal of automotive technology}, vol.~19, pp. 837--844, 2018.

\bibitem{suo2021trafficsim}
S.~Suo, S.~Regalado, S.~Casas, and R.~Urtasun, ``Trafficsim: Learning to
  simulate realistic multi-agent behaviors,'' in \emph{Proceedings of the
  IEEE/CVF Conference on Computer Vision and Pattern Recognition}, 2021, pp.
  10\,400--10\,409.

\bibitem{varadarajan2022multipath++}
B.~Varadarajan, A.~Hefny, A.~Srivastava, K.~S. Refaat, N.~Nayakanti,
  A.~Cornman, K.~Chen, B.~Douillard, C.~P. Lam, D.~Anguelov \emph{et~al.},
  ``Multipath++: Efficient information fusion and trajectory aggregation for
  behavior prediction,'' in \emph{2022 International Conference on Robotics and
  Automation (ICRA)}.\hskip 1em plus 0.5em minus 0.4em\relax IEEE, 2022, pp.
  7814--7821.

\bibitem{chang2022analyzing}
W.-J. Chang, Y.~Hu, C.~Li, W.~Zhan, and M.~Tomizuka, ``Analyzing and enhancing
  closed-loop stability in reactive simulation,'' in \emph{2022 IEEE 25th
  International Conference on Intelligent Transportation Systems (ITSC)}.\hskip
  1em plus 0.5em minus 0.4em\relax IEEE, 2022, pp. 3665--3672.

\bibitem{li2024residual}
C.~Li, C.~Tang, H.~Nishimura, J.~Mercat, M.~Tomizuka, and W.~Zhan, ``Residual
  q-learning: Offline and online policy customization without value,''
  \emph{Advances in Neural Information Processing Systems}, vol.~36, 2024.

\bibitem{zhan2017spatially}
W.~Zhan, J.~Chen, C.-Y. Chan, C.~Liu, and M.~Tomizuka, ``Spatially-partitioned
  environmental representation and planning architecture for on-road autonomous
  driving,'' in \emph{2017 IEEE Intelligent Vehicles Symposium (IV)}.\hskip 1em
  plus 0.5em minus 0.4em\relax IEEE, 2017, pp. 632--639.

\bibitem{li2022efficient}
C.~Li, T.~Trinh, L.~Wang, C.~Liu, M.~Tomizuka, and W.~Zhan, ``Efficient
  game-theoretic planning with prediction heuristic for socially-compliant
  autonomous driving,'' \emph{IEEE Robotics and Automation Letters}, vol.~7,
  no.~4, pp. 10\,248--10\,255, 2022.

\bibitem{schwarting2019social}
W.~Schwarting, A.~Pierson, J.~Alonso-Mora, S.~Karaman, and D.~Rus, ``Social
  behavior for autonomous vehicles,'' \emph{Proceedings of the National Academy
  of Sciences}, vol. 116, no.~50, pp. 24\,972--24\,978, 2019.

\bibitem{doi:10.1177/02783649231215371}
\BIBentryALTinterwordspacing
H.~Hu, D.~Isele, S.~Bae, and J.~F. Fisac, ``Active uncertainty reduction for
  safe and efficient interaction planning: A shielding-aware dual control
  approach,'' \emph{The International Journal of Robotics Research}, 2023.
  [Online]. Available: \url{https://doi.org/10.1177/02783649231215371}
\BIBentrySTDinterwordspacing

\bibitem{salzmann2020trajectron++}
T.~Salzmann, B.~Ivanovic, P.~Chakravarty, and M.~Pavone, ``Trajectron++:
  Dynamically-feasible trajectory forecasting with heterogeneous data,'' in
  \emph{Computer Vision--ECCV 2020: 16th European Conference, Glasgow, UK,
  August 23--28, 2020, Proceedings, Part XVIII 16}.\hskip 1em plus 0.5em minus
  0.4em\relax Springer, 2020, pp. 683--700.

\bibitem{gu2021densetnt}
J.~Gu, C.~Sun, and H.~Zhao, ``Densetnt: End-to-end trajectory prediction from
  dense goal sets,'' in \emph{Proceedings of the IEEE/CVF International
  Conference on Computer Vision}, 2021, pp. 15\,303--15\,312.

\bibitem{chang2023editing}
W.-J. Chang, C.~Tang, C.~Li, Y.~Hu, M.~Tomizuka, and W.~Zhan, ``Editing driver
  character: Socially-controllable behavior generation for interactive traffic
  simulation,'' \emph{arXiv preprint arXiv:2303.13830}, 2023.

\bibitem{wu2022toward}
T.~Wu, E.~Sachdeva, K.~Akash, X.~Wu, T.~Misu, and J.~Ortiz, ``Toward an
  adaptive situational awareness support system for urban driving,'' in
  \emph{2022 IEEE Intelligent Vehicles Symposium (IV)}.\hskip 1em plus 0.5em
  minus 0.4em\relax IEEE, 2022, pp. 1073--1080.

\end{thebibliography}

\end{document}